\documentclass[submission]{eptcs}
\providecommand{\event}{arXiv} 
\usepackage{breakurl}             
\usepackage{amsmath,amssymb,amsfonts}
\usepackage{algorithmic}
\usepackage[ruled,vlined]{algorithm2e}
\usepackage{tabularx}
\usepackage{multirow}
\usepackage{bm}
\usepackage{graphicx}
\usepackage{textcomp,amsthm}
\usepackage{xcolor,color}
\usepackage{multirow}
\usepackage[utf8]{inputenc}

\newcommand{\rset}{\mathbb{R}}

\newcommand{\R}{{\mathbb R}}

\newcommand{\be}{\begin{equation}}
\newcommand{\ee}{\end{equation}}
\newcommand{\ba}{\begin{array}}
\newcommand{\ea}{\end{array}}
\newcommand{\baa}{\left[\begin{array}}
\newcommand{\eaa}{\end{array}\right]}

\newcommand{\beqa}{\begin{eqnarray}}
\newcommand{\eeqa}{\end{eqnarray}}
\newcommand{\bt}{\begin{tabular}}
\newcommand{\et}{\end{tabular}}

\newcommand{\bi}{\begin{itemize}}
\newcommand{\ei}{\end{itemize}}
\newcommand{\bc}{\begin{center}}
\newcommand{\ec}{\end{center}}

\newtheorem{prop}{Proposition}
\newtheorem{remark}{Remark}

\newcounter{APctr} 

\newcounter{ABctr} 

\newcommand{\norm}[1]{\left\lVert#1\right\rVert}

\def\QED{\hfill \mbox{\rule[0pt]{1.5ex}{1.5ex}}}

\newcommand{\eref}[1]{(\ref{#1})}

\newlength{\algitab}
\title{Community-Level Anomaly Detection for Anti-Money Laundering
}

\author{}
\author{
    Andra Băltoiu
        \thanks{Acknowledgements: This work was partially supported by BRD Groupe Societe Generale through Data Science Research Fellowships of 2019 and partially by  a  grant  of  Romanian  Ministry  of  Research  and  Innovation  CCCDI-UEFISCDI.  project  no.  17PCCDI/2018.} 
    \institute{The Research Institute of the University of Bucharest (ICUB),\\University of Bucharest, Romania}
    \email{andra.baltoiu@fmi.unibuc.ro}
\and
    Andrei Pătrașcu~\footnotemark[1]
    \institute{Department of Computer Science,\\University of Bucharest, Romania}
    \email{andrei.pătrașcu@fmi.unibuc.ro}
\and
     Paul Irofti~\footnotemark[1]
     \institute{Department of Computer Science,\\University of Bucharest, Romania}
    \email{paul@irofti.net}
}

\def\titlerunning{Community-Level Anomaly Detection for Anti-Money Laundering}
\def\authorrunning{}

\begin{document}
\maketitle
\begin{abstract}

Anomaly detection in networks often boils down to identifying an underlying graph structure on which the abnormal occurrence rests on.
Financial fraud schemes are one such example, where more or less intricate schemes are employed in order to elude transaction security protocols.
We investigate the problem of learning graph structure representations using adaptations of dictionary learning aimed at encoding connectivity patterns.
In particular, we adapt dictionary learning strategies to the specificity of network topologies and propose new methods that impose Laplacian structure on the dictionaries themselves.
In one adaption we focus on classifying topologies by working directly on the graph Laplacian and cast the learning problem to accommodate its 2D structure.
We tackle the same problem by learning dictionaries which consist of vectorized atomic Laplacians, and provide a block coordinate descent scheme to solve the new dictionary learning formulation.
Imposing Laplacian structure on the dictionaries is also proposed in an adaptation of the Single Block Orthogonal learning method.
Results on synthetic graph datasets comprising different graph topologies confirm the potential of dictionaries to directly represent graph structure information.

\end{abstract}

\noindent \textbf{Keywords}: anomaly detection, dictionary learning, graph Laplacian classification, money laundering. \\

\
\section{Introduction}

The benefits of global inter-connectivity and 
the general increase of the quality of life 
led to the democratization of the general population's access to banking resources 
such as accounts, cards and cash machines.
Unfortunately,
this also led to ever more clever money laundering schemes 
that are becoming harder and harder to track
especially in the world of Big Data 
where finding anomalous behaviours or patterns in a pool of legitimate transactions
is like finding a needle in a haystack.

Recent years have shown great success in applying machine learning~(ML) techniques
when dealing with large chunks of data whose intrinsic structure is eluded.
Standard ML algorithms focus on the dominant trend,
but for anti-money laundering~(AML) we are interested
in their anomaly detection~(AD) variants
that have started to develop only in recent years \cite{akoglu2014graphbased, bolton2002, elliott2019anomaly}.
Here the focus is shifted towards the unfitted data.

Banking transactions between different entities (e.g. individuals, companies, banks, state institutions) can be represented as a directed weighted graph:
the nodes are the entities connected by directed edges
representing the transactions whose weight is given by various transaction and entity attributes (e.g. amount, currency, country of origin etc.).
Please note that describing all transactions within a time-frame (a month, a trimester, a year) leads to a very large graph.
It is common for AML techniques to look for known static patterns
within the existing transactions in order to identify possible frauds.
Thus the first problem that we focus on is identifying patterns,
or sub-graphs, in a given graph.
This task might seem daunting at first, and it is indeed NP-hard,
but we will use community detection and ML techniques to gain some traction.
Community detection is used to split the graph into manageable sub-graphs
and ML to perform pattern matching within these sub-graphs.
A second approach is to simply perform AD without any prior knowledge such as known patterns or other inside information from the bank institutions.
This is more challenging even with ML algorithms but, if successful,
it has the benefit of providing new insight into present money-laundering schemes.
Unlike normal ML tasks,
the key insight here is to over-fit on existing data in order to strongly reject anomalous behaviours.

Numerous signal processing applications are now being recast in order to include structural information, in particular information about the topology of the underlying network. These underlying graphs can have physical interpretations, as in the case of natural or man-made networks, or may be abstract structures, relevant to the data, as in the case of images. 
Either way, paying attention to the graph topology is proving fruitful in signal processing tasks \cite{Dong19_spm}.

Since many real networks pose scaling issues due to being prohibitively large to be handled as a whole, in practice they are broken down to sub-graphs or communities for further analysis.
Graph classification is therefore becoming a mandatory tool, both in itself and as a means to an end.

Dictionary learning (DL) methods are well suited for signal processing applications
such as compressed sensing, image denoising, inpainting and blind source separation.
In this article we use DL to construct and identify graph structures and their connections to the signals that they produce or support. 

Our proposed algorithmic schemes either directly work with graph structures or are intended for the more general problem of learning signals that lie on graphs. 
The first type is therefore appropriate for applications where the topologies are available, such as those that concern most distributed clusters with known network structure.
While the second regards underlying hidden graphs with possibly no physical relevance, but which rather encode dependencies or correlations.

Both approaches are suited for the task of identifying malicious transaction schemes relevant to money-laundering. 
Domain knowledge reveals several typical patterns employed in fraudulent conducts, such as clique-like or circular structures.
Our first proposal targets precisely this case, where the known graph structure is defining for the signal's membership to the anomaly class.
However, other financial crimes remain undetected and the question of exposing these schemes (and their corresponding graph structures) is an open one.
The latter proposal is suited for this task, as it indirectly infers structural information from a signal resting on a graph.

\subsection{The Dictionary Learning Problem}

The sparse representation model assumes that a given signal admits a representation in which only few elements, fewer that its original 
dimension, are nonzero. 
More precisely, with the aid of an overcomplete basis called a dictionary,
the signal can be written as

\begin{equation}
	\label{eq:dl_problem}
	\bm{Y} = \bm{D} \bm{X} +\bm{V},
\end{equation}
where $\bm{Y} \in \R^{m \times N}$ are the signals, $\bm{X} \in \R^{n \times N}$ is representation with $s$ nonzero elements, $\bm{D} \in \R^{m \times n}$ is the dictionary and $\bm{V}$ zero mean Gaussian noise.
The columns of the dictionary are called atoms and in order to avoid the ambiguity introduced by multiplication, they are normalized.

The dictionary learning problem consists in finding both the dictionary and the 
representation that best characterize the signal. 
The objective function therefore is 

\be
\begin{aligned}
& \underset{\bm{D}, \bm{X}}{\min}
& & \norm{\bm{Y}-\bm{DX}}_F^2 \\
& \text{s.t.}
& & \norm{\bm{X}_\ell}_{0} \leq s,\ 1 \le \ell \le N \\
& & & \norm{\bm{D}_j} = 1, \ 1 \le j \le n
\end{aligned}
\label{dict_learn}
\ee

The common practice is to solve the above problem in an alternate manner, that is by first fixing the dictionary and computing the representation, then fixing the representation and updating the dictionary. 
The former step is also known as sparse coding and is usually performed either by greedy methods that attempt to construct the sparse support \cite{PRK93omp}, or by relaxing the $l_0$ pseudo-norm in \eqref{dict_learn} into a convex formulation, such as by replacing it with $l_1$.
As for the latter step, it is routinely solved by coordinate or gradient descent methods. K-SVD \cite{AEB06}, a coordinate descent method, is a particularly
popular choice.
\subsection{Classification using Dictionary Learning}\label{sec:classDL}

The representative power of the dictionary can also be employed to classification tasks, provided some additional measures are taken, 
that specialize atoms to the characteristics of each class of signals. 
Intuitively, signals corresponding to the same class should have similar representations,
that (ideally) differ significantly from those of other classes. 

The formulation in \eqref{eq:dl_problem} suggests this discriminative property does not arise naturally, however it can be pursued without modifying the objective function, by training separate (sub)dictionaries with signals from each class \cite{Skrett06}.
Class estimation of a new signal can then be obtained by comparing its representation error with each subdictionary.
The core of this scheme, known as Sparse Representation-based Classifier,  was developed in \cite{Wright_09}, however in this initial formulation the dictionaries are not learned, rather they are a collection of class representative signals. 
SRC-like methods that also perform learning, as described previously, are known to the field.
Additional penalties can be included in order to limit the possibility of having similar atoms in different subdictionaries or, furthermore, to explicitly impose discriminative traits on the dictionary by penalizing the representation errors on each class~\cite{MairalDisc08}.

An altogether different approach is to indirectly force the desired properties in $\bm{D}$ by imposing that $\bm{X}$ both fulfills the signal reconstruction demands and exhibits class expressiveness.
This translates in coupling representation learning with classification.
Label Consistent K-SVD (LC-KSVD)~\cite{JLD11} is such an approach, that explicitly constrains the representation to be similar for signals within the same class and that moreover imposes each dictionary atom to become specialized for one class only.

\subsection{Dictionary Learning on Graphs}
Most common DL adaptations that handle graph structure learning build on sparse coding applications, such as~\cite{Zheng11_GraphReg}, that employ Laplacian learning as a smoothness-inducing factor.
These methods assume that if data sits on a graph, then observations in neighborhood nodes have similar representations, a property known as local invariance. 
Forcing the representations to include the data manifold structure is achieved by adding a Laplacian regularization term to the standard sparse coding or DL formulation.

Classification, too, can gain from this constraint on the codes, as intra-class similarities are also assumed to be reflected in the underlying graph structure.
In \cite{YankElad17}, the constraint of LC-KSVD that ties atoms to classes is replaced with a milder regularization term that ensure smoothness over the data manifold. 

The suitability of dictionaries to implicitly represent structure has been tested in works such as \cite{IS17_ifac}, which train the dictionary to discriminate between several types of faults in water networks.

\vspace{10pt}

\noindent \textbf{Notations}. Given matrix $\bm{A}$, we denote $\bm{A}_i$ the $i\textsuperscript{th}$ column of $\bm{A}$ and $\bm{A}^j$ the $j\textsuperscript{th}$ line of $\bm{A}$. Also, given vector $x$, we use notation $(x)_{\ell}$ for $\ell^{\text{th}}$-block of $x$. Given the closed convex set $\mathcal{X}$ and a point $z$, we denote with $\pi_{\mathcal{X}}(z)$ the orthogonal projection of $z$ onto $\mathcal{X}$. For a given function $f$, we denote $\nabla f(z)$ the gradient of $f$ at $z$ and with $\nabla_{x_i} f(z)$ the $i^{\text{th}}$-block of the gradient at $z$. For any integer $n$, we have $[n] =\{1, \cdots, n\}$. For any matrix $A \in \rset^{m \times n}$, the operation $\text{vec}(A)$ returns a vector of dimension $mn$ containing the vectorized columns of $A$. The all-ones vector is denoted by $\mathbf{1}$ and  the Kronecker product by $\otimes$.


\section{Laplacian-structured Dictionary Learning}

We first consider the problem in which the structure of the graph is the defining quality for the application at hand and consequently the crux of the classification task. 
Thus, in this section we design a learning algorithm based on structured dictionaries for computing efficient representations of Laplacian matrices. We assume that each matrix signal $\bm{y}_i \in \rset^{m \times m}$ has a Laplacian structure that we intend to capture using linear combinations of basic atomic Laplacians:  
\begin{align*}
    \bm{y}_i  \approx \bm{L}^{(1)}x_1 + \cdots  + \bm{L}^{(n)} x_n\qquad \text{where} \; \bm{x} \in \rset^{n}, \bm{L}^{(j)} \in \rset^{m \times m}, \; \forall  j \in [n].
\end{align*}
More compactly, we seek the following particular representation of $\bm{y}_i$:
\begin{align*}
    \bm{y}_i  \approx \bm{L} \cdot \bm{x} \otimes \mathbf{1},
\end{align*}
where $\bm{x} \in \rset^n$ and  $\bm{L} = [\bm{L}^{(1)} \cdots \bm{L}^{(n)}] \in \rset^{m \times mn}$.
In order to put this approximation problem in a more formal manner, we consider vectorized signals $y_i \in \rset^{m^2}$, the atoms $\bm{D}_i = vec(\bm{L}^{(i)})$ and derive the following constrained optimization problem with Laplacian-type constraints:
\begin{align}\label{firstpb_dl}
\min\limits_{\bm{D},\bm{X},\bm{L}} & \; \norm{\bm{Y} - \bm{D}\bm{X}}^2_F \\
\text{s.t.} \;\;\; & \bm{D}_i = \text{vec}(\bm{L}^{(i)}), \;\;  1\le i \le n, \nonumber \\
             &  \bm{L}^{(i)} \mathbf{1} = 0,\;  \bm{L}^{(i)}_{\ell j} \le 0, \; \forall \ell \neq j, \nonumber \\  
              &  \sum_j \bm{L}^{(i)}_{jj} = m. \nonumber 
\end{align}
Note that the slack variables $\bm{L}$ can be easily eliminated, but we keep them for a more elegant presentation. A typical approach in the literature, for solving the problem \eqref{firstpb_dl}, is the K-SVD family of algorithms, which proved empirically efficient when only orthogonality constraints are present; however, the extension of K-SVD scheme to the more complicated Laplacian constraints is highly nontrivial and we do not analyze this subject in our paper. Consequently, we will focus on alternating minimization and block coordinate descent schemes which are more appropriate for our current constrained setting (see \cite{Nes12,PatNec:15,PatNec:15ieee}).

\vspace{5pt}

\noindent The last linear equality constraint of \eqref{firstpb_dl} couples the lines of the Laplacian matrices and it is imposed with the purpose of avoiding the trivial solution (see \cite{YankElad17}). Since the problem scales with Laplacian dimension $m^2$ and data dimension $N$, simple algorithms with computationally cheap iteration are ideal in our high dimensional setting. Although, the block coordinate descent (BCD) methods are one of the most appropriate schemes at hand (\cite{Nes12,PatNec:15,PatNec:15ieee}), they require constraints separability which do not hold in our case due to the last equality. To remedy this issue, we transfer the last equality constraint in the objective through quadratic penalization, which will further avoid the trivial solution, and subsequently apply BCD inner schemes. 
Also, since we aim to obtain economic representations of the vectorized Laplacians, we will additionally impose $\ell_0$ sparsity constraints on the representations matrix $X$, obtaining the following problem:   
\begin{align}\label{complete_dllapl}
\min\limits_{\bm{D},\bm{X},\bm{L}} & \; \norm{\bm{Y} - \bm{D}\bm{X}}^2_F + \frac{\rho}{2}\sum\limits_{i = 1}^n\left( Tr(\bm{L}^{(i)}) - m\right)^2\\
\text{s.t.} \;\;\; & \bm{D}_i = \text{vec}(\bm{L}^{(i)}), \;\;  1\le i \le n, \nonumber\\
             &  \bm{L}^{(i)} \mathbf{1} = 0,\;  \bm{L}^{(i)}_{kj} \le 0, \; \forall k \neq j, \nonumber\\  
              & \norm{\bm{X}_i}_0 \le s, \nonumber
\end{align}
where $s$ is the prefixed number of nonzeros in each representation. 
Notice that, for high enough values of the penalty parameter $\rho>0$, the trace of each atomic Laplacian $\bm{L}^{(i)}$ approaches $m$. Since the resulted problem has nonconvex objective function and also nonconvex constraints, one of the most simple and efficient outer strategies in this case is the alternating minimization (AM) scheme, which, at each iteration, alternates the minimization over $\bm{D}$ and $\bm{X}$.

\begin{algorithm}
\SetKwComment{Comment}{}{}
\KwData{signals $\bm{Y} \in \R^{m^2 \times N}$,
initial dictionary $\bm{D}^0 \in \R^{m^2 \times n}$, $k = 0$
}
\BlankLine

\While {stopping criterion do not hold}
{
Compute $\bm{X}^{k}$ by solving \eqref{complete_dllapl} over $\bm{X}$, with $\bm{D} = \bm{D}^{k}$ fixed \\
Compute $\bm{D}^{k+1}$ by solving \eqref{complete_dllapl} over $\bm{D}$, with $\bm{X} = \bm{X}^k$ fixed \\
$k = k + 1$}

\caption{Alternating Minimization (AM)}
\label{alg:AM}
\end{algorithm}
\noindent Notice that, since none of the steps of AM are computable in closed-form, we will consider AM as an outer optimization scheme and provide further appropriate algorithms to solve the local problems from each step. 

\noindent \textbf{Computing the dictionary}. The optimization problem in $\bm{D}$ is convex, with a linear least squares type objective and linear constraints, which can be approached in principle with standard QP algorithms.
Given the representation matrix from previous iteration $\bm{X}^k$, the new dictionary is the solution of the following problem:
\begin{align}\label{dict_problem}
\bm{D}^{k} = \arg\min\limits_{\bm{D},\bm{L}} & \; \; f_{\rho}(\bm{D},\bm{L}) : = \norm{\bm{Y} - \bm{D}\bm{X}^k}^2_F + \frac{\rho}{2}\sum\limits_{i = 1}^n\left(Tr(\bm{L}^{(i)})- m\right)^2\\
\text{s.t.} \;\;\; & \bm{D}_i = \text{vec}(\bm{L}^{(i)}), \;\;  1\le i \le n, \nonumber\\
             &  \bm{L}^{(i)} \mathbf{1} = 0,\;  \bm{L}^{(i)}_{kj} \le 0, \; \forall k \neq j. \nonumber
\end{align}
Since this particular problem scales with $m^2$ and $N$, most of general QP iterative algorithms (e.g. gradient algorithms \cite{Nes83}, interior point methods)  require at each iteration at least $\mathcal{O}(m^2N)$ operations, which even for medium dimensions $(m,N)$ might be prohibitively high. 

\noindent Therefore, we further provide a BCD scheme with $\mathcal{O}(mn + m \log{(m)})$ complexity per iteration, which efficiently computes the dictionary even in high dimensions. The BCD algorithm uses the constraints separability of \eqref{dict_problem}(see \cite{Nes12}): each line of the Laplacian $\bm{L}^{(i)}$ is subjected to a simplex-type set (one linear equality and positivity constraints). Since the projection of simplex-type sets can be efficiently computed in $\mathcal{O}(m \log{m})$, then a projected coordinate gradient step can also have a fast iteration. Therefore, our scheme performs the following steps:
\begin{enumerate}
\item[$(i)$] Chooses randomly at each iteration an atom $\bm{D}_i$ and a $m-$size block $\bm{D}_{il}$ from the atom. This random chosen block represents a line from the vectorized Laplacian $\bm{L}^{(i)}$. 
\item[$(ii)$] Once the randomized selections are made, a projected coordinate gradient descent step is performed on the objective function.
\end{enumerate}
Formally, by eliminating the slack variables $\bm{L}$, recall that we obtain a new form of penalized objective function 
$$f_{\rho}(\bm{D}) := \norm{\bm{Y} - \bm{D} \bm{X}}^2_F + \frac{\rho}{2}\sum\limits_{i = 1}^n \left(\sum\limits_{j = 0}^{m-1} (\bm{D}_i)_{j(m+1)+1} - m\right)^2.$$ By further denoting $\bm{e_I} = vec(\bm{I}_m)$ and $\bm{R}_{(i)} = \sum\limits_{j \neq i} \bm{D}_j \bm{X}^j$, then we have the gradient on atom $\bm{D}_i$:
\begin{align*}
\nabla_{\bm{D}_i} f_{\rho}(\bm{D}) 
&= (\bm{D}_i\bm{X}^i - \bm{R}_{(i)})(\bm{X}^i)^T + \rho \bm{e_I} \left(\sum\limits_{j = 0}^{m-1} (\bm{D}_i)_{j(m+1)+1} - m \right)\\
&= \bm{D}_i\norm{\bm{X}^i}^2 - \bm{R}_{(i)}(\bm{X}^i)^T + \rho \bm{e_I} \left(\sum\limits_{j = 0}^{m-1} (\bm{D}_i)_{j(m+1)+1} - m \right).
\end{align*}
The gradient $ \nabla_{\bm{D}_i} f_{\rho}(\bm{D})$ is Lipschitz continuous in $\bm{D}$ and in order to determine the Lipschitz constants we make the following observations. Given two matrices $(\bm{D},\bm{\tilde{D}})$ which differ only on the $\ell-$th block of the $i$-th atom with difference vector $\bm{h} \in \rset^m$, then:  $$[\bm{\tilde{D}}_j]_{t} = 
\begin{cases}
[\bm{D}_j]_{t} + \bm{h} & \text{if} \;\; j = i, t = \ell \\
[\bm{D}_j]_{t} & \text{if} \;\; j \neq i \\
[\bm{D}_j]_{t} & \text{if} \;\; j = i, t \neq \ell.
\end{cases} $$
But observing that 
$$\norm{[\nabla_{\bm{D}_i} f(\bm{D})]_{\ell}    - [\nabla_{\bm{D}_i} f(\bm{\tilde{D}})]_{\ell} } \le (\norm{\bm{X}^i}^2 + \rho) \norm{\bm{h}} \quad \forall \bm{D} \in \rset^{m^2 \times n}, \bm{h} \in \rset^m,$$
then it is straightforward to estimate the Lipschitz constants as  $L_i = \norm{\bm{X}^i}^2 + \rho$. 

\noindent The traditional constant stepsize used for most of the first-order algorithms, including BCD schemes, is inverse proportional with the Lipschitz constant. Therefore, we will further use the derived Lipschitz constant estimate in the stepsize of $\ell-$th block-atom update in BGCD scheme as follows:
\begin{align*}
[\bm{D}_i^{k+1}]_{\ell} = [\bm{D}_i^{k}]_{\ell} - \frac{1}{L_{i}} [\nabla_{\bm{D}_i} f(\bm{D}^k,\bm{X}^k)]_{\ell}.
\end{align*}
Denoting $\mathcal{X}_{\ell} = \{d \in \rset^m: \bm{1}^Td = 0, d_{\ell} \ge 0, d_j \le 0, \forall j \neq \ell\}$, we provide below the complete BCGD iteration:
\begin{algorithm}
\SetKwComment{Comment}{}{}
\KwData{signals $\bm{Y} \in \R^{m^2 \times N}$,
initial dictionary $\bm{D}^0 \in \R^{m^2 \times n}$, representations $X \in \rset^{n \times N}$
}
\BlankLine

\While {stopping criterion do not hold}
{ Choose randomly $ 1\le i_k \le n, 1 \le \ell_k \le m$ \\
Compute gradient step: $ \bm{\mathcal{D}} = [\bm{D}_{i_k}^{k}]_{\ell_k} - \frac{1}{L_{i_k}} [\nabla_{\bm{D}_{i_k}} f(\bm{D}^k,\bm{X}^k)]_{\ell_k}$ \\
 Compute projection step: $
 [\bm{D}_j^{k+1}]_{t} = ,
 \begin{cases}
\pi_{\mathcal{X}_{\ell}}(\bm{\mathcal{D}}) & \text{if} \;\; j = i_k, t = \ell_k \\
[\bm{D}_j]_{t} & \text{if} \;\; j \neq i_k \\
[\bm{D}_j]_{t} & \text{if} \;\; j = i_k, t \neq \ell_k.
\end{cases}$ \\
 k = k + 1
 }
\caption{Block Coordinate Gradient Descent (BCGD)}
\label{alg:BCGD}
\end{algorithm}

\noindent The main computational effort is comprised in the gradient and projection steps. The first one can be straightforwardly estimated to take $\mathcal{O}(mn)$. However, if we consider that the matrix $X^k(X^k)^T$ is available, then the complexity can be further reduced. 
For the projection step we use Kiwiel's algorithms \cite{Kie07}, which requires $\mathcal{O}(m \log{(m)})$ operations.
Notice that, usually the stopping criterion is based on the diminishing gradients norm of the objective function $f$.  
Regarding the total computational complexity, sublinear and dimension-dependent convergence rates for multiple BCGD schemes have been provided in \cite{Nes12,PatNec:15,PatNec:15ieee} for convex and non-convex (sparse) optimization problems. 

\vspace{10pt}

\noindent \textbf{Computing the representation}. Secondly, the representation problem in $\bm{X}$ is a nonconvex $\ell_0$-constrained quadratic sparse representation problem which can be sub-optimally solved with standard schemes such as OMP.  
\begin{align}\label{repr_problem}
\bm{X}^{k+1}= \arg\min\limits_{X} & \; \norm{\bm{Y} - \bm{D}^k \bm{X}}^2_F \\
\text{s.t.} \;\;\; & \norm{\bm{X}_i}_0 \le s.\nonumber
\end{align}

\begin{remark}
Proper theoretical analysis of the overall computational complexity of the outer AM scheme, using the convergence rates of the inner BCGD algorithm, will be provided in a  future work. 
\end{remark}


\section{Separable Laplacian Classification} \label{sec:SLL}
\subsection{Preliminaries on Separable Dictionary Learning}
The DL framework can also handle multidimensional signals, however the standard 
formulation requires the signals to be vectorized.
This operation breaks the correlations between successive columns of the data matrix, be they spatial, as in the case of images, or temporal, as in multiple measurements applications. 
The alternative, developed in \cite{Hawe13_SepDL}, is to train one dictionary for each dimension of the signal, such that each captures the patterns occurring along it. 
Consider the case of 2D signals, $\bm{Y} \in \R^{m_1 \times m_2 \times N}$.
The DL problem can be expressed in terms of two dictionaries, 
$\bm{D}_1 \in \R^{m_1 \times n_1}$ and $\bm{D}_2 \in \R^{m_2 \times n_2}$, and 
the representation $\bm{X} \in \R^{n_1 \times n_2 \times N}$

\begin{equation}
	\label{eq:problem_2D}
	\bm{Y} = \bm{D}_1 \bm{X} \bm{D}_2^\top + \bm{V}
\end{equation} 

The model is equivalent to \eqref{eq:dl_problem} considering 
$\bm{D} = \bm{D}_2 \otimes \bm{D}_1$, where $\otimes$ stands for the Kronecker 
product. 
As such, \eqref{eq:problem_2D} can be plugged in the objective function \eqref{dict_learn}
to obtain the straightforward 2D adaption. Note that the normalizing constraint on the dictionary atoms now applies to both $\bm{D}_1$ and $\bm{D}_2$.

Both computing the representation and updating the dictionary can benefit from working with a structured, smaller problem. 
Several extensions of existing methods have been proposed for the two steps, with 
evident advantages in what complexity is concerned and similar, if not equal, performance 
compared to the standard algorithms. 
We briefly describe the ones used later in our simulations.

The sparse coding step can be efficiently computed via an adaptation of the OMP 
algorithm \cite{FangHuang12_2DOMP} that considers pairs of atoms from $\bm{D}_1$ 
and $\bm{D}_2$ when updating the support, resulting in a significant complexity reduction
of a factor of $m$ compared to applying standard OMP on the vectorized formulation.

As for the dictionary learning step, Pairwise Approximate K-SVD \cite{ID19_icassp} 
performs an alternate update of the two dictionaries. 
Keeping one dictionary fixed when renewing the atoms of the other ensures that 
competing pairs that share one atom do not hinder convergence.
The update itself is in the spirit of the original AK-SVD.

\subsection{Laplacian Structure}
Our next strategy is to exploit the two dimensional structure of a graph Laplacian in order to learn connectivity patterns that are specific to each class of graphs. 
The adaption of the classification method previously presented to the separable structure case proceeds as follows.


The training signals from each class are used separately to train one pair of dictionaries.
Classifying a new, test signal, will account to evaluating which pair of dictionaries is better at representing the signal.
This ability is assessed by computing the root mean square error, defined as 
$\text{RMSE} = \frac{1}{\sqrt{m_1 m_2 N}} \norm{\bm{Y} - \bm{D_1} \bm{X} \bm{D_2}^\top}_F$.
The smallest of the $c$ errors associated to each signal marks the estimated class. 
Algorithm \ref{alg:SepClassDL} resumes the steps above.

We abuse the superscript $(c)$ notation to indicate variables relating to class $c$.
Dictionaries  $\bm{D_1}^{(c)}$ and $\bm{D_2}^{(c)}$ are trained on signals that 
belong to $c$, $\bm{Y}^{(c)}$. 

\begin{algorithm}
\SetKwComment{Comment}{}{}
\KwData{train signals $\bm{Y}_{train} \in \R^{m_1 \times m_2 \times N_{train}}$,
 test signals $\bm{Y}_{test} \in \R^{m_1 \times m_2 \times N_{test}}$, \\
 \hspace{\algitab} train labels, test labels, sparsity level $s$, classes $C$, \\
  \hspace{\algitab} initial dictionaries $\bm{D_1}^{(c)} \in \R^{m_1 \times n_1}$,  
  $\bm{D_2}^{(c)} \in \R^{m_2 \times n_2}$
}
\KwResult{estimated test labels, $\bm{\hat{l}}$}
\BlankLine

\For {$c = 1:C$}{
	Train class dictionaries: $\bm{D_1}^{(c)}, \bm{D_2}^{(c)}$ using signals  $\bm{Y}_{train}^{(c)}$ and $s$ \\
	Compute the representation $\bm{X}^{(c)}$ of all test signals $\bm{Y}_{test}$ using previously learned $\bm{D_1}^{(c)}, \bm{D_2}^{(c)}$\\
	Compute representation errors: $\bm{\epsilon}_c = \norm{\bm{Y}_{test} - \bm{D_1}^{(c)} \bm{X}^{(c)} \bm{D_2}^{(c)\top} }_F^2 $\\
}
Classify test signals: $\bm{\hat{l}}_{test} = argmin (\bm{\epsilon}) $\\

\caption{Graph Laplacian Classification using Separable Dictionary Learning}
\label{alg:SepClassDL}
\end{algorithm}

\section{Graph Orthonormal Blocks Classification}
\subsection{Preliminaries on the Single Block Orthogonal method}
The standard DL problem 
designs dictionary $\bm{D}$ as a set of $n$ independent atoms,
but for certain classes of signals it has been shown \cite{Irofti15_composite}
that imposing structure on $\bm{D}$ and its atoms improve its representation power.
A common approach is to structure the dictionary as a union of orthonormal blocks \cite{Lessage05_uonb}\cite{Rusu13_sbo}.
The DL problem becomes
\be
\begin{aligned}
& \underset{\bm{Q}_1, \ldots, \bm{Q}_L, \bm{X}}{\min}
&& \norm{\bm{Y} - \left[\bm{Q}_1 \ \bm{Q}_2 \ \dots \ \bm{Q}_L \right]\bm{X}}^2_F \\
& \text{s.t.}
&& \|\bm{x}_i\|_0 \leq s, \ 1 \le i \le N \\
&&& \bm{Q_j}^T\bm{Q_j} = \bm{I}_m, \ 1 \le j \le L
\end{aligned}
\label{union_learn}
\ee
where each block of atoms $\bm{Q}$ represents a set of atoms,
forming an orthogonal basis,
that are optimized together.
It is common to instill these blocks with extra properties
that allow for faster or better approximation algorithms.

Single Block Orthogonal (SBO) algorithm \cite{Rusu13_sbo}
builds the dictionary as a union of $L$ orthogonal basis (or blocks) $\bm{Q}\in\R^{m\times m}$.
Given signal $\bm{y}$,
the representation stage identifies the basis that represents it best
and uses thresholding to impose sparsity.

\begin{remark}\label{rem:thresh}
Due to orthogonality,
the hard-thresholding operation of
canceling all but the $s$ absolute largest coefficients of
$\bm{x}=\bm{Q}^T\bm{y}$
is optimal.
\end{remark}

Thus,
the $s$-sparse representation $\bm{x}$ of a signal $\bm{y}$ 
using block $\bm{Q}$ can be simply implemented through the partial sorting function:
$\bm{x} = \text{SELECT}(\bm{Q}^T\bm{y}, s)$.

\begin{prop}~\cite[Chapter 7]{DL_book} \rm
\label{prop:Qalloc}
Given a union of $L$ orthogonal blocks,
the best basis $j$ to represent a given signal $\bm{y}$
is picked by computing the energy of the resulting representation coefficients
and selecting the block where the energy is highest.
More precisely, if
\be
\bm{x}^i = \text{SELECT}(\bm{Q}_i^T \bm{y}, s)
\ee
is the representation using block $\bm{Q}_i$,
and $E_{\bm{x}^i} = \norm{\bm{x}^i}^2$ is its energy, then the best orthogonal block is
given by
\be
j = \arg \underset{i = 1:L}{\max}(E_{\bm{x}^i}).
\label{sbo_alloc}
\ee
\QED
\end{prop}

During the dictionary refinement stage,
each basis $\bm{Q}_j$ is updated based on the $N_j$ signals using it for representation (in the same spirit as K-SVD).
The minimization of the representation error with such a dictionary is
called the orthogonal Procrustes problem.

\begin{prop}~\cite[Chapter 4]{DL_book} \rm
\label{prop:Qopt}
Given the matrices $\bm{Y}, \bm{X} \in \R^{m \times N_j}$,
if $\bm{Q}_j \in \R^{m \times m}$ is orthogonal, then the approximation error
$\norm{\bm{Y}-\bm{Q}_j\bm{X}}_F$ is minimized by
\be
\bm{Q}_j = \bm{VU}^T,
\label{orth_opt}
\ee
where the matrices $\bm{U}$, $\bm{V}$ are obtained from the singular value
decomposition
\be
\bm{XY}^T = \bm{U \Sigma V}^T.
\label{svd_XY}
\ee
Here $\bm{\Sigma}$ is diagonal, $\bm{U}$ and $\bm{V}$ are orthogonal
and all are $m \times m$.
\QED
\end{prop}

A special characteristic of the SBO algorithm is that it starts with an initial set of bases,
which it then expands during training.
Before adding a basis to the existing set,
a given percentage of the worst represented signals are collected in $\bm{W}$
and the new basis is initialized 
by performing a few rounds of training on $\bm{W}$.

\begin{remark}[Improvements via Parallelism]\label{rem:paral}
The special dictionary structure of SBO
and its effects on the representation and training stages
make it a good candidate for parallelization.
Indeed,
this is demonstrated by the GPU implementation from \cite{Irofti15_sbo}
where P-SBO is proposed (P stands for parallel).
P-SBO expands the set of bases
by more than one at a time
which,
besides improving parallelism,
has been shown to also reduce the representation error.
\end{remark}

\begin{remark}[Advantages over AK-SVD]\label{rem:timeadv}
Due to the simplicity of the representation stage,
$O(m^2)$ for representation and $O(m)$ for thresholding,
the main computation demand of SBO is the dictionary training stage,
whose complexity is driven by the final number of bases in the dictionary
and the number of rounds necessary to train them.
On the other hand,
OMP is usually the algorithm of choice
when performing representation with unstructured dictionaries
due to its speed and representation power.
However,
it is also the main computational bottleneck of the K-SVD family,
having a complexity of $O(ms(Lm+s^2))$,
which makes AK-SVD pale in comparison with SBO in terms of execution time for both DL and signal representation 
(see the execution times from Tables 2 and 3 in ~\cite{Irofti15_sbo}).
\end{remark}

\subsection{Laplacian Blocks}

Imposing Laplacian structure on the dictionary can also be achieved by adapting
the SBO algorithm presented previously.

Indeed,
during training,
we initialize each orthogonal basis with a fixed an orthogonalized Laplacian matrix.
The SBO training rounds performed on each basis will further adapt and refine this original laplacian to better fit the signals using the current basis.

The end goal is to classify signals that are known to lie on a graph whose structure is unknown.
This is achieved at the end of the learning process by
employing a SRC-like scheme for handling the classification task,
that can be now performed by simply using the basis allocation scheme described in Proposition \ref{prop:Qalloc}.

As such,
our application performs a separate SBO instance for each class specific signal training set.
The key is to initialize the bases with the orthogonalized true Laplacian of the corresponding class
and then we can proceed with the standard P-SBO.
The resulting union of bases of each class $Q^{(c)}$ are collected together and classification consists in applying \eref{sbo_alloc} on the set of all bases
such that the chosen basis indicates the signal class.


\section{Results}

The aim of the following two synthetic experiments is to show how transferring structural information from the graphs to the dictionary atoms themselves improves the performance in classification tasks when compared to standard DL methods that are oblivious to the nature of the signals.
While the first considers the case where the signals directly express network topologies, the second deals 
with more generic signals that rest on a graph.

While the proposed solutions are all suitable for the general classification problem, in the subsequent tests the focus is on anomaly detection. 
In real-world financial data, fraudulent network topologies occur rarely compared to legitimate ones and the following synthetic experiments are constructed in order to assess the ability of the methods in representing such isolated events.

\subsection{Anomalous Graph Laplacians}

Our first experiment is designed to test the ability of our proposed methods
to distinguish between different network structures.
As such, we test our Separable Laplacian Learning and the Laplacian-structured method on a synthetic graph 2-class classification dataset, constructed with the application of anomaly detection in mind.
All signals are generated to have the same network topology, while on a small number of signals we implant an anomaly, namely a structurally distinct subgraph.
The normal graphs are constructed using a stochastic block model with 8 modules, strong intra-module connectivity (controlled by the diagonal dominance of the probability matrix) and inter-module probability of 0.05. Each graph has $n_1 = 50$ nodes.

Anomalies are smaller, $n_2 = 10$ nodes, Watts-Strogatz graphs with mean degree $k = 4$ and rewiring probability $\beta = 0.2$. 
The configuration leads to a network with prominent circular structure.
All weight values, regardless of whether the edges are in the normal or anomalous part of the graph, are distributed normally in $[0, 100]$. 
We construct $N_1 = 5000$ normal graphs and $N_2 = 500$ anomalies and compute
their Laplacian matrices.
We only consider the case of undirected graphs, however the solutions can be used for directed graphs, as well. 

We test the ability of two of our approaches in learning to classify these 
Laplacians and compare with the classic SRC-like scheme, where, as described 
in Section \ref{sec:classDL}, we train different dictionaries on the signals of each class. 
Moreover, we compare the DL methods with One-Class Support Vector Machine (OC-SVM) \cite{TiaMah:18}, a reliable unsupervised anomaly detection method, that has been successfully employed for anti-money laundering tasks \cite{JunJia:05}. 
OC-SVM takes the true anomaly ratio ($10\%$) as input parameter in order to derive the decision bound between the normal signals and the outliers. 

In order to suit the classical DL and OC-SVM algorithms, we vectorize the Laplacian matrices. Vectorization is also needed in our proposed Laplacian-structured Dictionary Learning method. Our Separable Laplacian Classification adaptation, 
on the other hand, works directly on the matrices.

The appropriate sparsity level is not given before hand and optimal values are 
usually determined through empirical tests.
There is however the typical value of $\sqrt m$, which is known to yield good results in some applications. 
An adaptation of this popular choice to the problem of Laplacian learning asks for setting $s$ to be roughly the square root of the number of edge connections in the graph. However, 
since this value differs slightly in different network realizations, for consistency reasons we set $s = 30$ in all experiments.
All the algorithms were applied on $80\%$ of the signals. The remaining $20\%$ are used to test the classification performance.

Classification accuracy (i.e. percentage of correctly labeled graphs) results presented in Table \ref{tbl:res} show that imposing or exploiting structure leads to increased performance. For compactness, we have use the "L-structured DL" abbreviation to denote our Laplacian-structured dictionary learning method; "Separable L-Class" to denote our Separable Laplacian Classification algorithm and refer to the standard SRC-like classification method as "DL Classification".

Results show that when the DL model incorporates known information on the signals (i.e. their Laplacian structure), the solution leads to a more accurate identification of anomalous graphs, compared to the "blind" SRC-like method and OCSVM.
Money laundering schemes most often entail unnatural connections between the nodes of the transactions graph, such as circular patterns, addressed in our synthetic experiment. The performance of the proposed methods is encouraging with respect to identifying other anomalous network structures as well.

\begin{table} \label{tbl:res}
\caption{Learning Graph Laplacians. Classification Accuracy (\%).}
	\makebox[\textwidth][c]{
		\begin{tabular}{|c|c|c|c|}
			\hline
			L-structured DL & Separable L-Class & DL Classification & OC-SVM \\
			\hline
			91.31 & 90.64 & 89.55 & 81.1\\
			\hline
		\end{tabular}
	}
\end{table}  
\subsection{Anomalous Signals on Graphs}
In a second set of experiments we test our adaptation of SBO to the 
problem of classifying signals that sit on different graph topologies.
It is common practice in synthetic DL experiments to generate the 
data as a linear combination of $s$ atoms of a known random 
dictionary.
In our case, we also require the signals to rest on a known graph with
Laplacian $\bm{L}$. 
We follow the principle described in \cite{Yankelevsky16_dualgraph} 
that ensures a proper coupling between the generating Laplacian 
and dictionary and construct the signals of each class using the 
following scheme.

Starting with a random initial $\bm{D_0}$, we obtain our true dictionary
as $\bm{D} = (\lambda \bm{I} + \bm{L})^{-1} \bm{D_0}$, where $\bm{L}$ is the graph Laplacian of the corresponding class.
We use the same topologies as in the previous experiment.
The parameter $\lambda$ controls the smoothness of the dictionary atoms. 
While atom smoothness is not crucial to our application, the choice of
the parameter will be reflected in how well the signals adhere to the
underlying graph structure.
As per the above mentioned reference, we set $\lambda=5$.
Following the construction of $\bm{D}$, we generate the signals using
$s=4$ random atoms and add Gaussian noise of level $SNR=20$.
We set $N_1 = 6000$ for the normal class and $N_2 = 600$ for the 
anomalies.

\begin{figure}[h]
\centering
\includegraphics[width=0.6\linewidth]{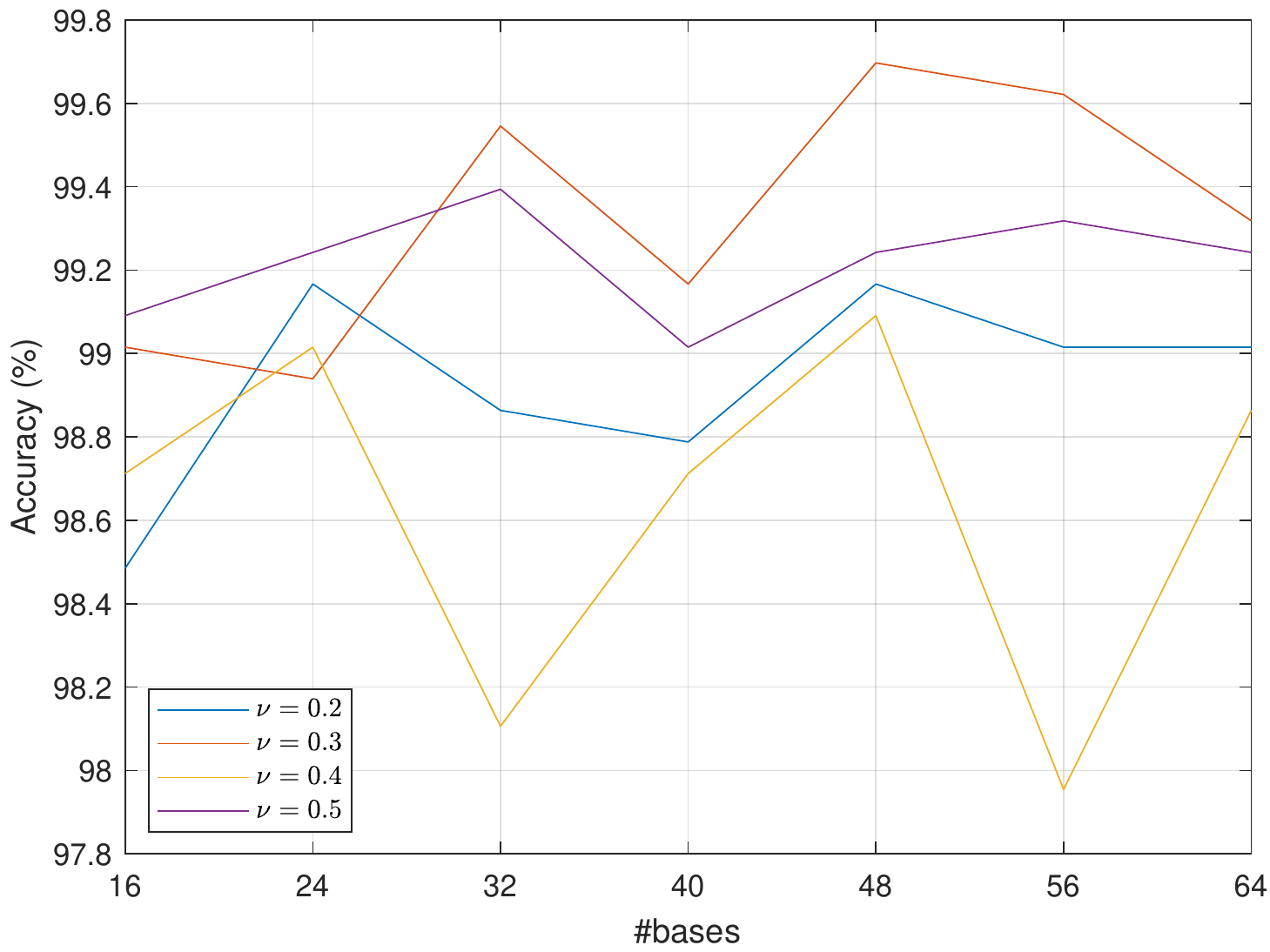}
\caption{Influence of SBO parameters on classification performance}
\label{fig:exp2}
\end{figure}.

As previously, $80\%$ of the signals are used for training, while the 
remaining for testing.
To initialize the union of orthogonal blocks, we orthogonalize the 
true generating Laplacians.
These initial 2 blocks are subsequently refined using $N/8$ signals 
of the corresponding class.
Figure \ref{fig:exp2} shows the influence that the number of blocks and fraction $\nu$ of badly represented signals used for basis construction have on performance.
We perform 7 rounds of base refinement and learn 6 bases in parallel (see Remark \ref{rem:paral}).

Best results, $\textbf{99.70}\%$ classification accuracy, are obtained when working with $48$ bases for each class and retraining on $30\%$ of the signals.
Applying the SRC-like classification on the dataset yields $\textbf{99.77}\%$.
The small difference is however compensated by the definite complexity 
advantage of SBO over SRC~\cite{Irofti15_sbo}, as detailed in Remark \ref{rem:timeadv}.

\section{Conclusions}

We proposed three dictionary learning methods for exploiting the structural information of graphs in order to improve the network classification task.
The solutions are aimed at identifying anomalous structures, targeting but not limited to anti-money laundering applications. 
When working directly with the structure of the graphs,
our method of imposing Laplacian structure on the dictionary atoms has yielded the better results compared to the standard dictionary classification algorithm and to OCSVM.
Our adaptation of the separable dictionary learning problem, which takes into account vicinity patterns in 2D data also constitutes a better alternative to the classical solution, which is oblivious to the underlying structure, as well as to OCSVM.
As for the more general problem of signals that lie on graphs, our adaptation of a block orthogonal algorithm, that 
imposes a Laplacian-like structure on the dictionary has yielded similar performance compared to the classic dictionary classification method, 
however with known computational advantages.
We have tested the suitability of these methods for anomaly detection applications by testing on a dataset with relevant class imbalance.

\bibliographystyle{eptcs}
\bibliography{from}

\end{document}